\documentclass{article}
\usepackage{graphicx}
\usepackage{amsmath}
\usepackage{amssymb}
\usepackage{hyperref}
\usepackage{cite}
\usepackage{authblk}

\title{\textbf{Visualizing Uncertainty in Image-Guided Surgery -- a Review }}

\author[ ]{\parbox{\textwidth}{\centering \small Mahsa Geshvadi $^{1,2,3}$}}

\affil[1]{\small University of Massachusetts Boston}
\affil[2]{\small Brigham and Women's Hospital}
\affil[3]{\small Harvard Medical School}

\date{}
\begin{document}

\maketitle

During tumor resection surgery, surgeons rely on neuronavigation to locate tumors and other critical structures in the brain. Most neuronavigation is based on preoperative images, such as MRI and ultrasound, to navigate through the brain. Neuronavigation acts like GPS for the brain, guiding neurosurgeons during the procedure. However, brain shift, a dynamic deformation caused by factors such as osmotic concentration, fluid levels, and tissue resection, can invalidate the preoperative images and introduce registration uncertainty. Considering and effectively visualizing this uncertainty has the potential to help surgeons trust the navigation again.

Uncertainty has been studied in various domains since the 19th century \cite{Pang1996Approaches}. Considering uncertainty requires two essential components: 1) quantifying uncertainty; and 2) conveying the quantified values to the observer. There has been growing interest in both of these research areas during the past few decades. 

\textbf{Uncertainty Visualization}
Studies have been done to provide an overview of existing methods for visualizing or communicating uncertainty, offering different perspectives on uncertainty visualization. Weiskopf et al. ~\cite{Weiskopf2022Uncertainty} presents a general overview of uncertainty visualization techniques and some examples of applying these techniques in bioinformatics. They focus on exploring layouts for advanced uncertainty visualization. Gillmann et al. ~\cite{Gillmann} provides an overview of uncertainty in medical imaging. Their paper explores three sources of uncertainty: image acquisition, transformation, and visualization. It focuses mainly on using colors to convey uncertainty and introduces available visualization techniques for various imaging modalities, including magnetic resonance imaging (MRI) and ultrasound.
Similarly, Pang et al. \cite{Pang1996Approaches} propose a classification system for uncertainty visualization techniques with five characteristics: value, location, data extent, visualization extent, and axes mapping. Ristowsky et al. \cite{Gordan2014} introduce a taxonomy for uncertainty in medical imaging. They categorize different types of uncertainty based on their factors such as spatial locations where the uncertainty is discrete or continuous, 2D or 3D, etc. Padilla et al.\cite{padilla_kay_hullman_2020} provide a variety of application-specific approaches.
They conclude that uncertainty visualization has no one-size-fits-all solution and emphasize considering design choices. The paper highlights the complexity of uncertainty visualization and the importance of empirical testing to ensure the effectiveness of visualizations.
These studies guided our understanding of factors to consider in visualizing uncertainty and underscored the complexity of uncertainty visualization.

Brodlie et al. \cite{Brodlie2012} categorize uncertainty visualization algorithms into three distinct classes: dense, sparse, and embedded. Dense visualizations display data at every point within a domain, whereas sparse visualizations focus on extracting and highlighting significant features, such as contour lines. The embedded approach involves placing visualizations into a higher-dimensional display space. For dense visualizations, they introduced two distinct approaches to visualizing uncertainty in contouring: value uncertainty and positional uncertainty. Value uncertainty visualizes the uncertainty along the mean contour line, often using techniques like uncertainty ribbons, where the contour line's thickness or color indicates the uncertainty level. Positional uncertainty visualizes the range of possible contour lines for a given threshold, illustrating the variability in the independent variable space, commonly visualized with methods like spaghetti plots.

Several prior approaches for providing uncertainty visualization use glyphs \cite{george2014Overview, Johnson2003AnextStep, Pang1996Approaches } and color overlays\cite{Berge2015Real-time, Gillmann}.
Less conventional approaches include the use of Augmented Reality \cite{Shamir2011Trajectory} and animation \cite{Djurcilov2002, Lundström2007Uncerr, Charles1997Vis}.
Grigoryan and Rheingans proposed a method for visualizing the probabilistic uncertainty of the shape of a 3D surface model \cite{Grigoryan2002Prob}. Osorio and Brodlie developed methods to visualize uncertainty during contouring \cite{Kim2008Contouring}.
In other contexts
Kay et al. \cite{Kay2016When} integrate interactive control in a system to help users understand uncertainty in predicting bus arrival times.
Greisi et al. \cite{Greis2018Uncertainty} also implement an application that gives users some control over uncertainty visualization, albeit in sensor measurement and for visualization of discrepant information. Simpson et al. \cite{simpson2006} perform a study on osteoid osteoma excision surgical task. The authors use volume rendering to visualize the uncertainty, which allows the path distribution to be viewed as a 3D volume or 2D cross section. The authors conducted a user study to evaluate the effectiveness of the visualization method. The task mimicked the excision of a deep bone tumor. The results showed that the visualization method resulted in a statistically significant reduction in the number of attempts required to localize a target.


For volumetric data uncertainty visualization studies, Athawale et al. \cite{Athawale2021} present a nonparametric statistical framework for visualizing uncertainty in volumetric data using direct volume rendering (DVR). They employ quantile interpolation to integrate nonparametric probability density functions (PDFs), thus enhancing the precision of uncertainty visualizations. Their approach includes extending to 2D transfer functions (TFs) for better classification and utilizing a quartile view to highlight reconstruction variability across different data quantiles. This method demonstrates improved accuracy over traditional parametric models. Djurcilov et al. \cite{Djurcilov2002} use two approaches to visualize uncertainty in volumetric data. Their inline DVR method incorporates uncertainty directly into the rendering process using transfer functions, mapping data to color and uncertainty to opacity. Their post-processing techniques modify volume-rendered images to indicate uncertainty by adding speckles, depth-shaded holes, noise, or texture. We explored a similar approach, which adds Gaussian blurring or noise that is correlated with uncertainty. Liu et al. \cite{liu2012} propose Gaussian Mixture Model-based volume visualization, which uses per-voxel Gaussian mixture models to represent and render volumetric data with uncertainty. This method reduces data storage and utilizes the GPU to provide real-time rendering. It visualizes uncertainty through animated flickering and detailed still frames. However, its visual complexity makes it unsuitable for the operating room. Potter et al. \cite{Potter2013}  use entropy as a summary statistic for categorizing data, such as each voxel of brain tissue, into one of 11 types and visualize uncertainty by highlighting high-entropy regions in white. This method highlights regions where the assignment to a particular category is uncertain, as indicated by higher entropy values. However, this approach changes the data representation and can result in the loss of important information, making it unsuitable for our context.

For neurosurgery, Frisken et al. \cite{Frisken2022Incorporating} propose a method that composes uncertainty contributed by image segmentation and brain shift into a single risk volume and conveys this risk to the surgeon during surgical planning using soft boundaries and volume rendering. That paper focuses on modeling and visualizing uncertainty during path planning, while our focus is on developing visualization methods that are effective in the complex environment of the operating room. 
Similarly, Diepenbrock et al. \cite{Diepenbrock2011Visualization} propose a method to give neurosurgeons a quick overview of the most important structures at risk during surgical planning. Their methods employed an intuitive red-blue color mapping in which nearby at-risk critical structures are rendered red. They introduce a workflow for path planning and target the planning phase. However, their approach does not account for registration uncertainty, which is the central aim of our study.

\section*{Conclusions}

Uncertainty in image-guided surgery can be introduced by many sources, including imaging, image processing, tracking, modeling, measurement in the operating room, etc. Improving precision in image-guided surgery may depend on surgeons being able to understand and visualize this uncertainty. Although uncertainty visualization is recognized as an important focus in visualization research, there has been relatively little work in this area for medical applications and particularly for image-guided surgery. We believe this is an important area of future work.

\bibliographystyle{IEEEtran}
\bibliography{main}

\end{document}